\documentclass{article}
\usepackage{ijcai21}

\usepackage{times}
\usepackage{soul}
\usepackage{url}
\usepackage[hidelinks]{hyperref}
\usepackage[utf8]{inputenc}
\usepackage[small]{caption}
\usepackage{graphicx}
\usepackage{amssymb}
\usepackage{amsmath}
\usepackage{amsthm}
\usepackage{booktabs}
\usepackage{algorithm}
\usepackage{algorithmic}
\usepackage{CJKutf8}
\usepackage{color}

\title{Few-shot Neural Human Performance Rendering from Sparse RGBD Videos}

\author{
Anqi Pang$^{1,2,3}$
\and
Xin Chen$^{1,2,3}$\footnote{The corresponding authors. \\Anqi Pang and Xin Chen are contributed equally.}\and
Haimin Luo$^{1,2,3}$\and
Minye Wu$^{1,2,3}$\and
Jingyi Yu$^1$\and
Lan Xu$^1$\footnote{ Corresponding author.}\\

\affiliations
$^1$ Shanghai Engineering Research Center of Intelligent Vision and Imaging, School of Information Science and Technology, ShanghaiTech University\\
$^2$ Shanghai Institute of Microsystem and Information Technology, Chinese Academy of Sciences\\
$^3$ University of Chinese Academy of Sciences\\

\emails
\{pangaq, chenxin2, luohm, wumy, yujingyi, xulan1\}@shanghaitech.edu.cn
}

\begin{document}

\maketitle

\begin{abstract}
Recent neural rendering approaches for human activities achieve remarkable view synthesis results, but still rely on dense input views or dense training with all the capture frames, leading to deployment difficulty and inefficient training overload. 
However, existing advances will be ill-posed if the input is both spatially and temporally sparse.
To fill this gap, in this paper we propose a few-shot neural human rendering approach (FNHR) from only sparse RGBD inputs, 
which exploits the temporal and spatial redundancy to generate photo-realistic free-view output of human activities. 
Our FNHR is trained only on the key-frames which expand the motion manifold in the input sequences.
We introduce a two-branch neural blending to combine the neural point render and classical graphics texturing pipeline, which integrates reliable observations over sparse key-frames.
Furthermore, we adopt a patch-based adversarial training process to make use of the local redundancy and avoids over-fitting to the key-frames, which generates fine-detailed rendering results.
Extensive experiments demonstrate the effectiveness of our approach to generate high-quality free view-point results for challenging human performances under the sparse setting.
\end{abstract}

\newcommand{\itemnudege}{\vspace{-.05in}}
\newcommand{\eqnbreak}{\par\vspace{-1.5\baselineskip}}
\newcommand{\vnudge}{\vspace{-.15in}}

\definecolor{Red}{cmyk}{0,1,1,0}
\definecolor{Green}{cmyk}{1,0,1,0}
\definecolor{Cyan}{cmyk}{1,0,0,0}
\definecolor{Purple}{cmyk}{0.45,0.86,0,0}
\definecolor{Rosolic}{cmyk}{0.00,1.00,0.50,0}
\definecolor{Blue}{cmyk}{1.00,1.00,0.00,0}
\definecolor{BlueViolet}{cmyk}{0.86,0.91,0,0.04}
\definecolor{NavyBlue}{cmyk}{0.94,0.54,0,0}

\newcommand{\other}[1]{{\color{Rosolic}  Other: #1}}
\newcommand{\tmp}[1]{{\color{Purple}  Tmp: #1}}
\newcommand{\yu}[1]{{\color{Red} \bf \em Yu: #1}}

\newcommand{\xu}[1]{{\color{BlueViolet} xu: #1}}
\newcommand{\ma}[1]{{\color{BlueViolet} ma: #1}}
\newcommand{\wei}[1]{{\color{Green}  Wei: #1}}
\newcommand{\chen}[1]{{\color{Red} chen: #1}}
\newcommand{\jin}[1]{{\color{Blue} jin: #1}}
\newcommand{\revised}[1]{{\color{BlueViolet} #1}}
\newcommand{\pang}[1]{{\color{BlueViolet} #1}}

\newcommand{\weiNote}[1]{{\color{red}  Wei: #1}}
\newcommand{\xuNote}[1]{{\color{red} xu: #1}}
\newcommand{\maNote}[1]{{\color{red} ma: #1}}
\newcommand{\chenNote}[1]{{\color{red} chen: #1}}
\newcommand{\pangNote}[1]{{\color{red} pang: #1}}

\newtheorem{thm}{Theorem}
\newtheorem{cor}[thm]{Corollary}
\newtheorem{lem}[thm]{Lemma}
\newtheorem{prop}[thm]{Proposition}
\newtheorem{defn}[thm]{Definition}
\newtheorem{rem}[thm]{Remark}
\let\vec=\mathbf
\let\set=\mathcal
\let\mat=\mathbf

\newcommand{\mypara}[1]{\paragraph*{#1.}}
\newcommand{\bv}[1]{\mathbf{#1}}
\newcommand{\todo}[1]{\hl{#1}}
\newcommand{\bi}[1]{\mathbf{#1}}

\newcommand{\myparagraph}[1]{\vspace{0.1em}\noindent\textbf{#1}}

\begin{CJK}{UTF8}{gbsn}
    \begin{figure}[t]
	\centering
	\includegraphics[width=\linewidth]{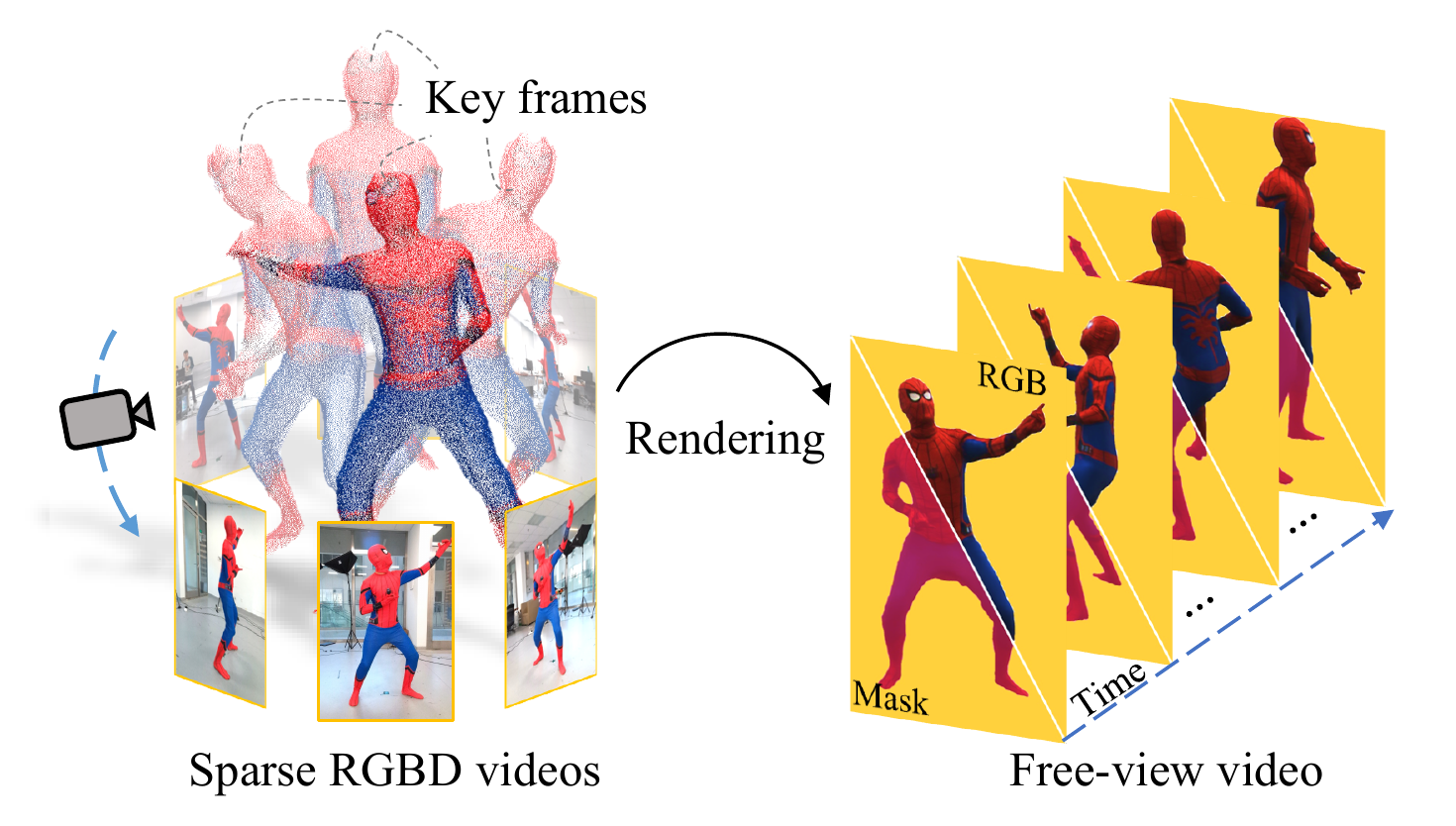}
	\caption{Our few-shot neural human rendering (FNHR) achieves photo-realistic free-view results from only six sparse RGBD inputs.}
	\label{fig:teaser}
\end{figure}

\section{Introduction}
The rise of virtual and augmented reality (VR and AR) to present information in an immersive way has increased the demand of the 4D (3D spatial plus 1D time) content generation. 
Further reconstructing human activities and providing photo-realistic rendering from a free viewpoint evolves as a cutting-edge yet bottleneck technique.

The early high-end volumetric solutions~\cite{motion2fusion,TotalCapture,collet2015high} rely on multi-view dome-based setup to achieve high-fidelity reconstruction and rendering of human activities in novel views but are expensive and difficult to be deployed.
The recent low-end approaches~\cite{UnstructureLan,robustfusion} have enabled light-weight and template-less performance reconstruction by leveraging the RGBD sensors and modern GPUs but are still restricted by the limited mesh resolution and suffer from the uncanny texturing output.

The recent neural rendering techniques~\cite{Wu_2020_CVPR,NeuralVolumes,nerf} bring huge potential for photo-realistic novel view synthesis and get rid of the heavy reliance on the reconstruction accuracy.
However, for dynamic scene modeling, these approaches rely on dense spatial capture views and dense temporal training with all the capture frames, leading to deployment difficulty and inefficient training overload.
On the other hand, few-shot or key-frame based strategy has been widely studied for human motion analysis~\cite{mustafa20164d}, revealing the temporal and spatial redundancy of human activities.
However, the literature on few-shot neural human performance rendering remains sparse.
Several recent works~\cite{shysheya2019textured} generated realistic neural avatars even based on key-frame inputs, but they rely on human
body model and can hardly handle topology changes, leading to severe visual artifacts for complex performance.

In this paper, we attack these challenges and present \textit{FNHR} -- the first \textbf{F}ew-shot \textbf{N}eural \textbf{H}uman performance \textbf{R}endering approach using six sparse RGBD cameras surrounding the performer (see Fig.~\ref{fig:teaser}).
Our approach generates photo-realistic texture of challenging human activities in novel views, whilst exploring the spatially and temporally sparse capture setup.
Generating such a human free-viewpoint video by training on only spatially and temporally sparse input in an end-to-end manner is non-trivial.
To this end, our key idea is to explore effective neural render design to encode the spatial, temporal, and local similarities across all the inputs, besides utilizing the inherent global information from our multi-view setting.
We first formulate the key-frame selection as a pose-guided clustering problem to generate key-frames which expand the motion manifold in the input sequences.
Then, based on these key-frames with coarse geometry proxy, a novel two-branch neural rendering scheme is proposed to integrates reliable observations over sparse key-frames, which consists of a neural point renderer and a classical graphics texturing renderer in a data-driven fashion.
Finally, we introduce a patch-based training process in an adversarial manner to make use of the local redundancy, which not only avoids over-fitting to the key-frames, but also generates fine-detailed photo-realistic texturing results.
To summarize, our main technical contributions include:
\begin{itemize}
	\item We present the first few-shot neural human rendering approach, which can generate photo-realistic free-view results from only sparse RGBD inputs, achieving significant superiority to existing state-of-the-art.
	
	\item We propose a two-branch hybrid neural rendering design to integrate reliable observations over the sparse key-frames generated via an effective pose-guide clustering process. 
	
	\item We introduce a novel patch-wise training scheme in an adversarial manner to exploit the local similarities and provide fine-detailed and photo-realistic texture results. 
	
\end{itemize}

    \section{Related Work}
\paragraph{Human Performance Capture.}
Markerless human performance capture techniques have been widely adopted to achieve human free-viewpoint video or reconstruct the geometry. 
The high-end solutions~\cite{motion2fusion,TotalCapture,chen2019tightcap} require studio-setup with the dense view of cameras and a controlled imaging environment to generate high-fidelity reconstruction and high-quality surface motion, which are expensive and difficult to deploy. 
The recent low-end approaches~\cite{Xiang_2019_CVPR,chen2021sportscap} enable light-weight performance capture under the single-view setup. 
However, these methods require a naked human model or pre-scanned template. 
Recent method~\cite{UnstructureLan,robustfusion} enable light-weight and template-less performance reconstruction using RGBD cameras, but they still suffer from the limited mesh resolution leading to uncanny texturing output. 
Comparably, our approach enables photo-realistic human free-viewpoint video generation using only spatially and temporally sparse input.

\paragraph{Neural Rendering. }
Recent work have made significant process on 3D scene modeling and photo-realistic novel view synthesis via differentiable neural rendering manner based on various data representations, such as point clouds~\cite{Wu_2020_CVPR,aliev2019neural}, voxels~\cite{lombardi2019neural}, texture meshes~\cite{thies2019deferred} or implicit functions~\cite{park2019deepsdf,nerf,suo2021neuralhumanfvv}. 
These methods bring huge potential for photo-realistic novel view synthesis and get rid of the heavy reliance on reconstruction accuracy.

For neural rendering of dynamic scenes,  Neural Volumes~\cite{NeuralVolumes} adopts a VAE network to transform input images into a 3D volume representation and can generalize to novel viewpoints. 
NHR~\cite{Wu_2020_CVPR} models and renders dynamic scenes through embedding spatial features with sparse dynamic point clouds. 
Recent work~\cite{park2020deformable,li2020neural,xian2020space,tretschk2020non} extend the approach NeRF~\cite{nerf} using neural radiance field into the dynamic setting. 
They decompose the task into learning a spatial mapping from a canonical scene to the current scene at each time step and regressing the canonical radiance field. 
However, for all the methods above, dense spatial capture views or dense temporal training with all the capture frames are required for high fidelity novel view synthesis, leading to deployment difficulty and inefficient training overload.
On the other hand, recent works~\cite{shysheya2019textured}  generate realistic neural avatars even based on key-frame inputs, but they rely on human body model and can hardly handle topology changes, leading to severe visual artifacts for complex performance. 

Comparably, our approach explores the spatially and temporally sparse capture setup and generates photo-realistic texture of challenging human activities in novel views.

    \begin{figure*}[t]
	\centering
	\includegraphics[width=0.95\linewidth]{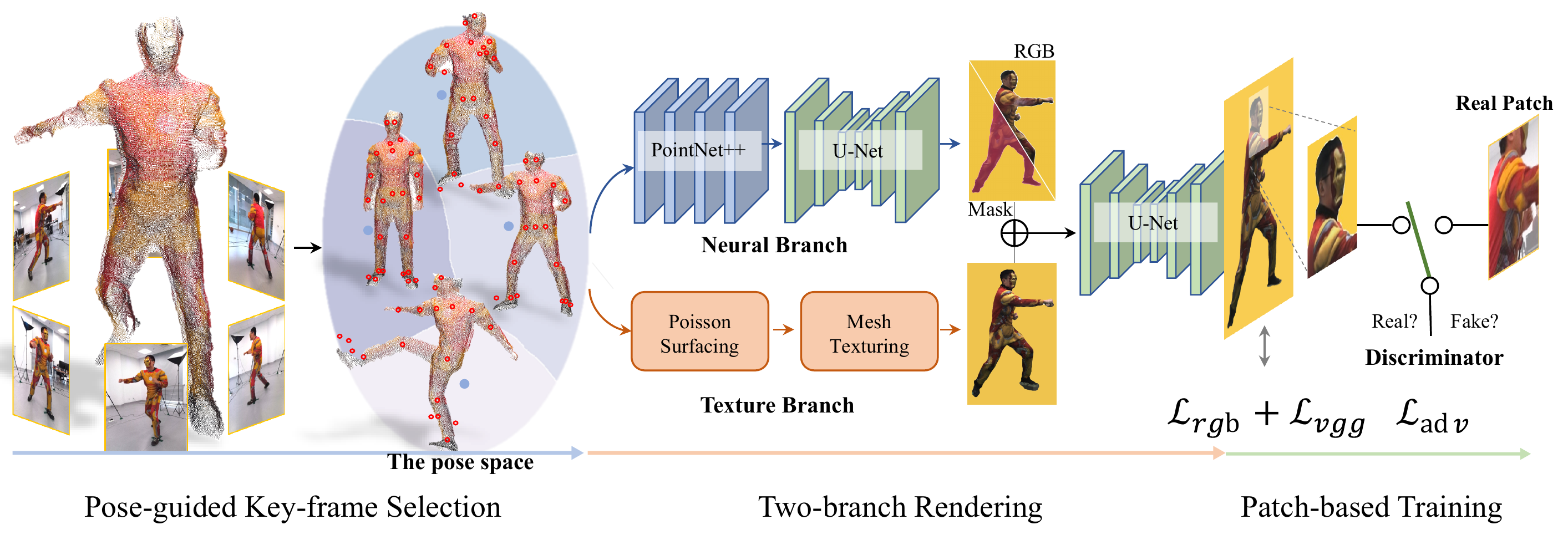}
	\caption{The pipeline of FNHR. Given the video inputs from six RGBD cameras surrounding the performer, our approach consists of {key-frame selection} (Sec. \ref{sec:keyframe}), {two-branch rendering} (Sec. \ref{sec:dualfusion}), and {patch-based training} (Sec. \ref{sec:patchedloss}) to generate free-view rendering results.}
	\vspace{-10pt}
	\label{fig:pipeline}
\end{figure*}

\section{Methods}
As illustrated in Fig.~\ref{fig:pipeline}, our FNHR explores neural human performance rendering under the few-shot and sparse-view setting to generate photo-realistic free-view results of human activities.
During training, the beauty of our approach lies in its light-weight reliance on only spatial and temporal key-frames to encode the appearance information of the whole human motion sequence, which breaks the deployment difficulty and inefficient training overload of previous methods.
To this end, we generate the key-frames to expand the motion manifold based on pose-guided clustering (Sec.~\ref{sec:keyframe}). 
Then, a two-branch neural rendering scheme is proposed to take a hybrid path between the recent neural point renderer and classical graphics pipeline (Sec.~\ref{sec:dualfusion}).
Our scheme extracts both the implicit and explicit appearance similarities of the performance between frames to overcome the spatial and temporal shortage of capture viewpoints. 
We also propose a patch-based adversarial re-renderer in FNHR to utilize the local redundancy, which avoids over-fitting to the key-frames and generates photo-realistic texturing details (Sec.~\ref{sec:patchedloss}).
Our approach takes only 6 RGBD streams from stereo cameras or Azure Kinect sensors surrounding the performer as input, and the human foreground mask is extracted using DeepLab v3.

\subsection{Pose-guided Key-frame Selection}\label{sec:keyframe}
Here, we introduce an effective scheme to select the representative key-frames to encode the information of the whole motion sequence, so as to avoid the heavy training process due to spatially and temporally superfluous training data.
%
%
Specifically, we first apply the OpenPose~\cite{OpenPose} followed by the triangulation on the six RGBD images to obtain a 3D human pose of the frame.
For triangulation, we optimize a global 3D skeleton with the depth to lift it from 2D to 3D. The per-joint confidences from all six views are utilized to filter out the wrong estimation of occlude joints.

Let $\mathcal{J}_t = \{ \mathbf{J}_t^1, \mathbf{J}_t^2,...,\mathbf{J}_t^V\}$ denote the predicted 3D pose of frame $t$, where $V=25$ is the number of body joints, while each element $\mathbf{J}_t^i=[x_i,y_i,z_i],i\in[1,V]$ corresponds to the 3D position of the body joint. 
Note that all these detected 3d poses are normalized into the same coordinates, and their root joints are aligned at the origin. 

Next, the key-frames selection is formulated as a pose-guided clustering problem, which ensures these key-frames to be representative in the motion manifold so that our neural rendering networks can be generalized to the entire sequence.
Thus, we conduct K-means in the pose space, the unsupervised clustering algorithm, with $k$ cluster centers, and each cluster center represents the 3D pose of a key-frame. 
Since each 3D pose $\mathcal{J}_t$ is a set of 3D joint positions, we define the following distance function to measure the difference between two poses $\mathcal{J}_x$ and $\mathcal{J}_y$ of various timestamps:
\begin{equation}
\label{equ:distance}
dist(\mathcal{J}_x, \mathcal{J}_y) = \sum_{i=1}^{V}\| \mathbf{J}_x^i - \mathbf{J}_y^i\|_2.
\end{equation}
Then, we calculate the numerical mean pose of each cluster and assign the frame with the nearest 3d pose such numerical central pose using the same distance function defined in Eqn.~\ref{equ:distance}.
The poses of these assigned frames are set to be the new cluster centers to ensure that the cluster centers are always located on exactly the frames from the sequence. 

After several clustering iterations until convergence, the key-frames of a sequence are selected from all the cluster centers. 
For multiple people scenarios, we concatenate the 3D human poses under the same timestamps and extend the Eqn.~\ref{equ:distance} to accumulate the differences of corresponding joint pairs. 
In our neural renderer training, we set $k$ to be 20 for a typical motion sequence with about 500 frames, leading to $4\%$ sparsity of capture view sampling.

\subsection{Two-branch Rendering} \label{sec:dualfusion}
Here we introduce a novel neural render design to encode the self-similarities across the spatially and temporally sparse key-frames.
Our key observation is that existing dynamic neural point renderers like \cite{Wu_2020_CVPR} and the graphics texturing pipeline are complementary to each other under our challenging sparse setting.  
The former one leads to photo-realistic texture in the input views but suffers from strong artifacts in between, while the latter one provides
spatially consistent rendering but the result suffers from reconstructed geometry error.
Thus, we propose a two-branch neural renderer to take a hybrid path between the recent neural point renderer and classical graphics pipeline, which utilizes both the implicit neural features and explicit appearance features to integrate reliable observations over sparse key-frames.

\vspace{2.5mm}\noindent{\bf Graphics Texturing Branch.}
To provide consistent rendering as a good anchor under our sparse setting, for the $t$-th frame, we use a similar fuse strategy to DynamicFusion to fuse the six depth images into a texture mesh $\mathbf{P}_t$ via Poisson reconstruction.
Then, classic graphics texturing mapping with mosaicing is adopted to generate a textured mesh and corresponding rendering image  $\mathbf{I}_\mathrm{tex}$.
Note that $\mathbf{I}_\mathrm{tex}$ suffers from texturing artifacts due to sparse-view reconstruction, but it preserves view coherent information and serves as an appearance prior for the following neural branch.

\paragraph{Neural Point Renderer Branch.}
We further adopt a neural point renderer to implicitly encode the appearance similarities between sparse key-frames in a self-supervised manner.
Similar to the previous dynamic neural point rendering approach~\cite{Wu_2020_CVPR}, we adopt a share-weighted PointNet++ on all the fused textured point clouds to extract point features and then the renderer splats point features into pixel coordinates on the target image plane with depth order to form the feature map. 
To avoid over-fitting to the key-frames due to our sparse setting, we randomly drop out $20\%$ points before feeding the point clouds into PointNet++.
Then, a modified U-Net with the gated convolution is applied on the feature map to generate the texture output with a foreground mask. 
Let $\psi_\mathrm{Neural}$ denote our neural renderer as follows:
\begin{equation}
\mathbf{I}_\mathrm{Neural}, \mathbf{M}_\mathrm{Neural} = \psi_{\mathrm{Neural}}(\mathbf{P}_t, \mathbf{K}, \mathbf{T}),
\label{eq:rendering}
\end{equation}
where $\mathbf{K}$ and $\mathbf{T}$ are the intrinsic and extrinsic matrices of the target view.
$\mathbf{I}_\mathrm{Neural}$ and $\mathbf{M}_\mathrm{Neural}$ are rendered color image and foreground mask, encoding the inherent appearance information from the key-frames into the render view.

\paragraph{Two-branch Blending.}
Finally, another U-Net based network $\psi_{\mathrm{fuse}}$ is adopted to fuse the above neural rendering result $\mathbf{I}_\mathrm{Neural}$ and the appearance prior $\mathbf{I}_\mathrm{tex}$ from our two branches, so as to obtain the final texture output as follows:
\begin{equation}
\mathbf{I}_{*} = \psi_{\mathrm{fuse}}(\mathbf{I}_\mathrm{Neural}, \mathbf{I}_\mathrm{tex}),
\label{eq:fusion}
\end{equation}
where $\mathbf{I}_\mathrm{Neural}$ and $\mathbf{I}_\mathrm{tex}$ are concatenated and fed into fuse net. 
Our blending scheme jointly utilizes both the implicit and explicit appearance similarities to overcome the spatial and temporal shortage of capture viewpoints.

\begin{figure}[t]
	\centering
	\includegraphics[width=\linewidth]{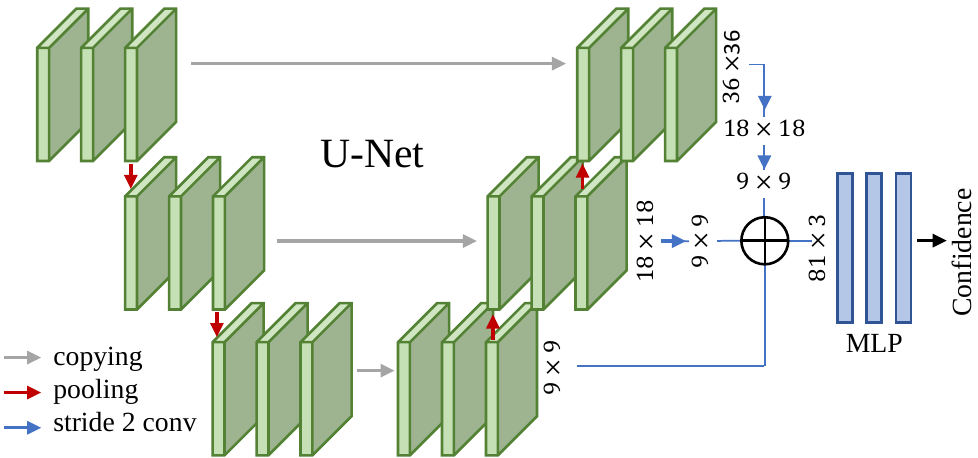}
	\caption{The network architecture of our multi-scale discriminator in patch-based adversarial training. }
	\label{fig:discriminator}
\end{figure}

\subsection{Patch-based Adversarial Training} \label{sec:patchedloss}
To handle the lack of diversity of training samples in our few-shot and sparse-view setting, we introduce a novel patch-based adversarial training scheme.
Our scheme improves the generalization of our neural render and the fine-grained details in the texture output significantly.
To this end, we randomly sample rectangular $36\times36$ patches from $\mathbf{I}_*$ and assign a discriminator network $D$ to predict if a patch comes from the real image or not.
Such patch-wise strategy narrows the scope of the real image distribution down and provides sufficiently diverse training samples to avoid overfitting to our small number of key-frames. 
Moreover, we mask background pixels of $\mathbf{I}_*$ out using the foreground masks $\mathbf{M}$ and define a patch as valid if the patch has $10\%$ of pixels belongs to foreground objects in our adversarial training.

\vspace{2.5mm}\noindent{\bf Multi-Scale Discriminator Network}. 
To distinguish the patches effectively, we design a multi-scale discriminator network as illustrated in Fig.~\ref{fig:discriminator}. 
Our discriminator $D$ adopts a 3-level U-Net to extract multi-scale features as well as a multiple layer perceptron (MLP) to aggregate features from different levels for final classification.
Specifically, at each level in the upsampling stream of U-Net, we use fully convolutional layers with a stride of 2 to downsample the feature map into a unified size of $9\times9$, and then flatten the feature map to a one dimension feature vector. 
All feature vectors from three levels are concatenated together as the input of the MLP. 
The MLP has three layers followed by ReLU activation with widths of 256, 128, 1, respectively. 
Such multi-scale features enhance the discriminative sensing ability to multi-scale details for our discriminator $D$. 
Besides, our two-branch renderer serves as the generator where the Pointnet++ and U-Net in $\psi_\mathrm{Neural}$ share the same architecture from \cite{Wu_2020_CVPR} while the U-Net in $\psi_{\mathrm{fuse}}$ has 5 level convolutional layers with skip connections similar to \cite{ronneberger2015u}.

\vspace{2.5mm}\noindent{\bf Training Details}. 
To enable few-shot and sparse-view neural human rendering, we need to train the fusion net $\psi_{\mathrm{fuse}}$ and discriminator $D$ simultaneously. But before that, we firstly bootstrap the neural point renderer branch on key-frame data set with 10 training epochs. 

Next, we exploit the following loss functions to supervise the training of the fusion net a and the discriminator. As described above, we apply patch-based training using the adversarial losses:
\begin{equation}
\begin{aligned}
\mathcal{L}_{adv_D}&= \frac{1}{\mid\mathcal{B}_{r}\mid}\sum_{x\in\mathcal{B}_{r}}(1 - D(x))^{2} +\frac{1}{\mid\mathcal{B}_{*}\mid}\sum_{y\in\mathcal{B}_{*}} D(y)^{2} \\
\mathcal{L}_{adv_G}&= -\frac{1}{\mid\mathcal{B}_{*}\mid}\sum_{y\in\mathcal{B}_{*}} D(y)^{2}, \\
\end{aligned}
\label{eq:ganloss}
\end{equation}
where $\mathcal{B}_{r}$ and $\mathcal{B}_{*}$ are the set of valid patches from real images and rendering results respectively; $\mathcal{L}_{adv_D}$ is only for updating the discriminator, and $\mathcal{L}_{adv_G}$ is only for updating the renderer; $\mathcal{L}_{adv}=\mathcal{L}_{adv_D}+\mathcal{L}_{adv_G}$. Here we use the L2 norm for better training stability in practice. Then gradients of $\mathbf{I}_*$ are accumulated from random sampled patches'. 

In addition to the patch-based adversarial loss $\mathcal{L}_{adv}$, we also apply L1 loss $\mathcal{L}_{rgb}$ and perceptual loss $\mathcal{L}_{vgg}$ on the output of fusion net as:
\begin{equation}
\begin{aligned}
\mathcal{L}_{rgb}&= \frac{1}{n}\sum_{i=1}^n\|\mathbf{I}_i-\mathbf{I}_i^*\|_1, \\
\mathcal{L}_{vgg}&= \frac{1}{n}\sum_{i=1}^n\|\Phi_\mathrm{VGG}(\mathbf{I}_i)-\Phi_\mathrm{VGG}(\mathbf{I}_i^*)\|_2,
\end{aligned}
\label{eq:L1VGG}
\end{equation}
where $n$ is the number of one batch samples; $\mathbf{I}$ is the masked ground truth image; $\Phi_\mathrm{VGG}(\cdot)$ extracts feature maps of input from the 2th and 4th layer of the VGG-19 network which is pretrained on ImageNet.

We linearly combine these three losses and have the total loss:
\begin{equation}
\begin{aligned}
\mathcal{L} = \lambda_{1} \mathcal{L}_{adv}+\lambda_{2} \mathcal{L}_{rgb}+\lambda_{3} \mathcal{L}_{vgg},
\end{aligned}
\label{eq:totalloss}
\end{equation}
where $\lambda_{1}$, $\lambda_{2}$ and $\lambda_{3}$ are weights of importance. We set them to 0.3, 5, and 0.7 respectively in our experiments. We use the Adam optimizer with a learning rate of 0.0002 to optimize the network parameters. The batch size is 4, and the number of one batch sample is 40.

We also augment our training data with random translation, random scaling, and random rotation. These transformations are applied on 2D images and camera parameters. Benefiting from these loss functions and training strategies, our FNHR can synthesize photo-realistic free-view video under the few-shot and sparse-view setting. 

    \begin{figure*}[t]
	\centering
	\includegraphics[width=0.95\linewidth]{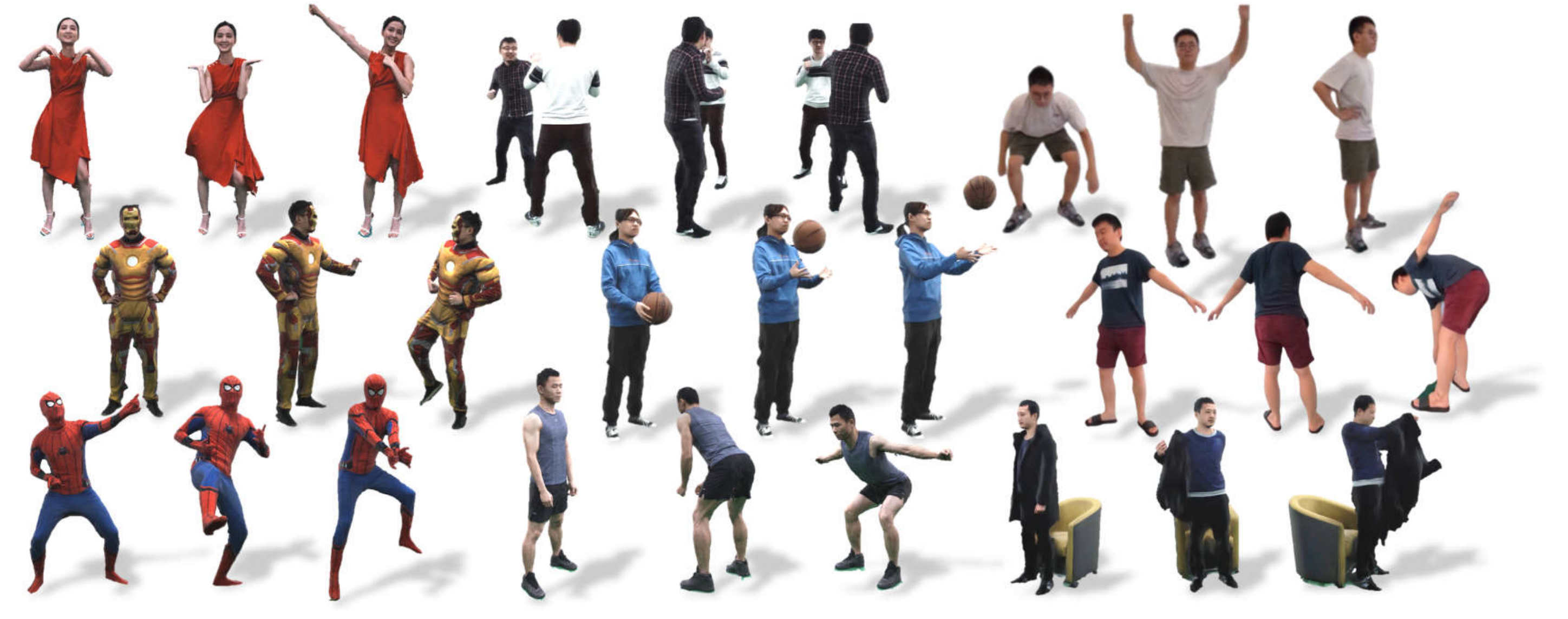}

	\caption{The photo-realistic free-view rendering texture results of the proposed few-shot neural rendering approach.}

	\label{fig:gallery}
\end{figure*}

\section{Experiment}
In this section, we evaluate our FNHR approach on a variety of challenging scenarios. 
%
We run our experiments on a PC with 2.2 GHz Intel Xeon 4210 CPU 64GB RAM, and Nvidia TITAN RTX GPU. It takes 541 ms to render a 720 $\times$ 1280 frame.
As demonstrated in Fig.~\ref{fig:gallery} our approach generates high-quality texture results under the few-shot setting and even handles human-object or multi-human interaction scenarios with topology changes, such as playing basketball, removing clothes, or fighting.

\begin{figure}[t]
    \centering
    \includegraphics[width=\linewidth]{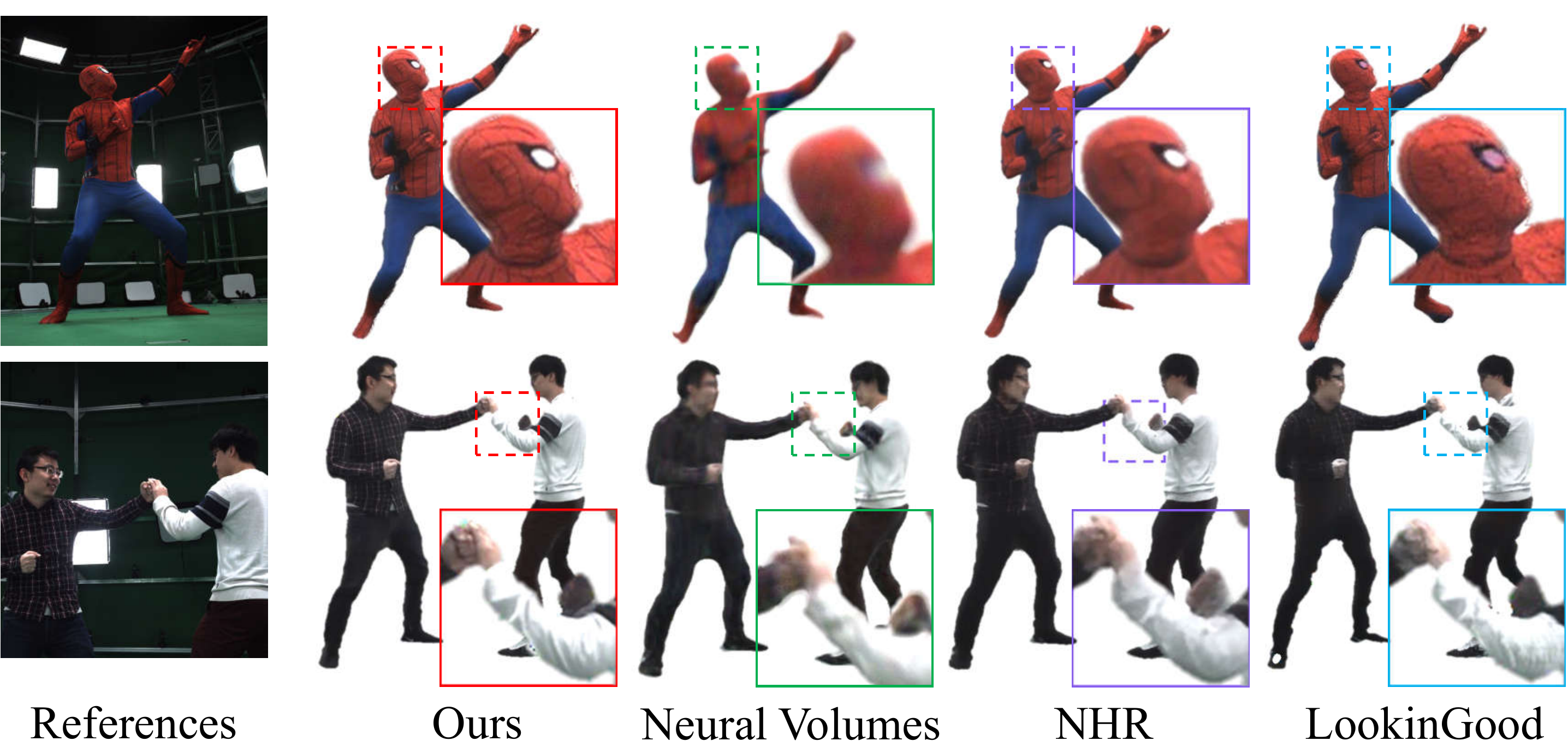}
    \caption{Qualitative comparison. Our approach generates more photo-realistic texture details than other methods.}

    \label{fig:comparsion}
\end{figure}

\begin{figure}[t]
    \centering
    \includegraphics[width=\linewidth]{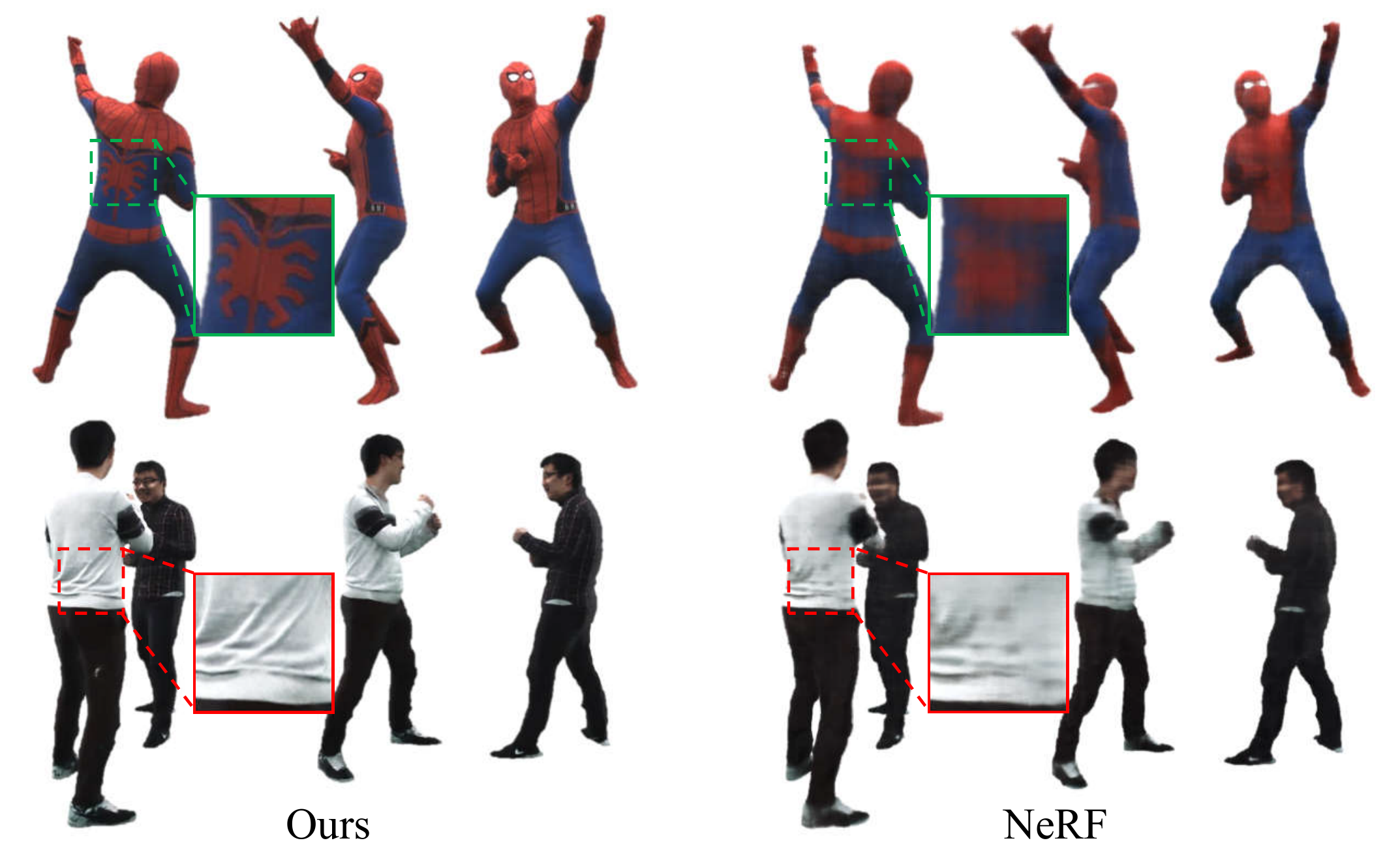}

    \caption{Qualitative comparison against NeRF. Using few-shot training data, our method achieves more sharp free-view rendering.}

    \label{fig:compare_nerf}
\end{figure}

\subsection{Comparison}
To the best of our knowledge, our approach is the first few-shot neural human rendering approach.
Therefore, we compare to existing dynamic neural rendering methods, including the voxel-based \textbf{Neural Volumes}~\cite{NeuralVolumes}, point-based \textbf{NHR}~\cite{Wu_2020_CVPR}, mesh-based \textbf{LookinGood}~\cite{LookinGood} using the same sparse training data for fair comparison.
As shown in Fig.~\ref{fig:comparsion}, all the other methods with various data representation suffer from uncanny texture results due to the shortage of training viewpoints.
In contrast, our approach achieves significantly better rendering results in terms of sharpness and realism, under the challenging sparse setting.
Then, we compare to the implicit method \textbf{NeRF}~\cite{nerf} based on the neural radiance field. 
Since NeRF only handles static scene, we use the six RGB images of the target frame as input and use the fused geometry from depths as the prior of initial density during the training of NeRF for a fair comparison.
As illustrated in Fig.\ref{fig:compare_nerf}, our approach reconstruct more sharp texture ouptut than NeRF under the few-shot setting.
Moreover, NeRF takes 2 hours for one frame training, leading to 800 hours for a sequence with frames, while our approach takes about 8 hours for training this sequence, achieving about $100\times$ speedup.

For quantitative comparison, we adopt the peak signal-to-noise ratio (\textbf{PSNR}), structural similarity index (\textbf{SSIM}), the L1 loss (\textbf{Photometric Error}) and the mean squared error (\textbf{MSE}) as metrics similar to previous methods.
Note that all the quantitative results are calculated in the reference captured views.
As shown in Tab.~\ref{tab:Comparison}, our approach consistently outperforms the other baselines in terms of all these metrics above, illustrating the effectiveness of our approach to handle the sparse setting and provide high-quality rendering results.

\subsection{Evaluation}
\begin{figure}[t]
	\centering
	\includegraphics[width=\linewidth]{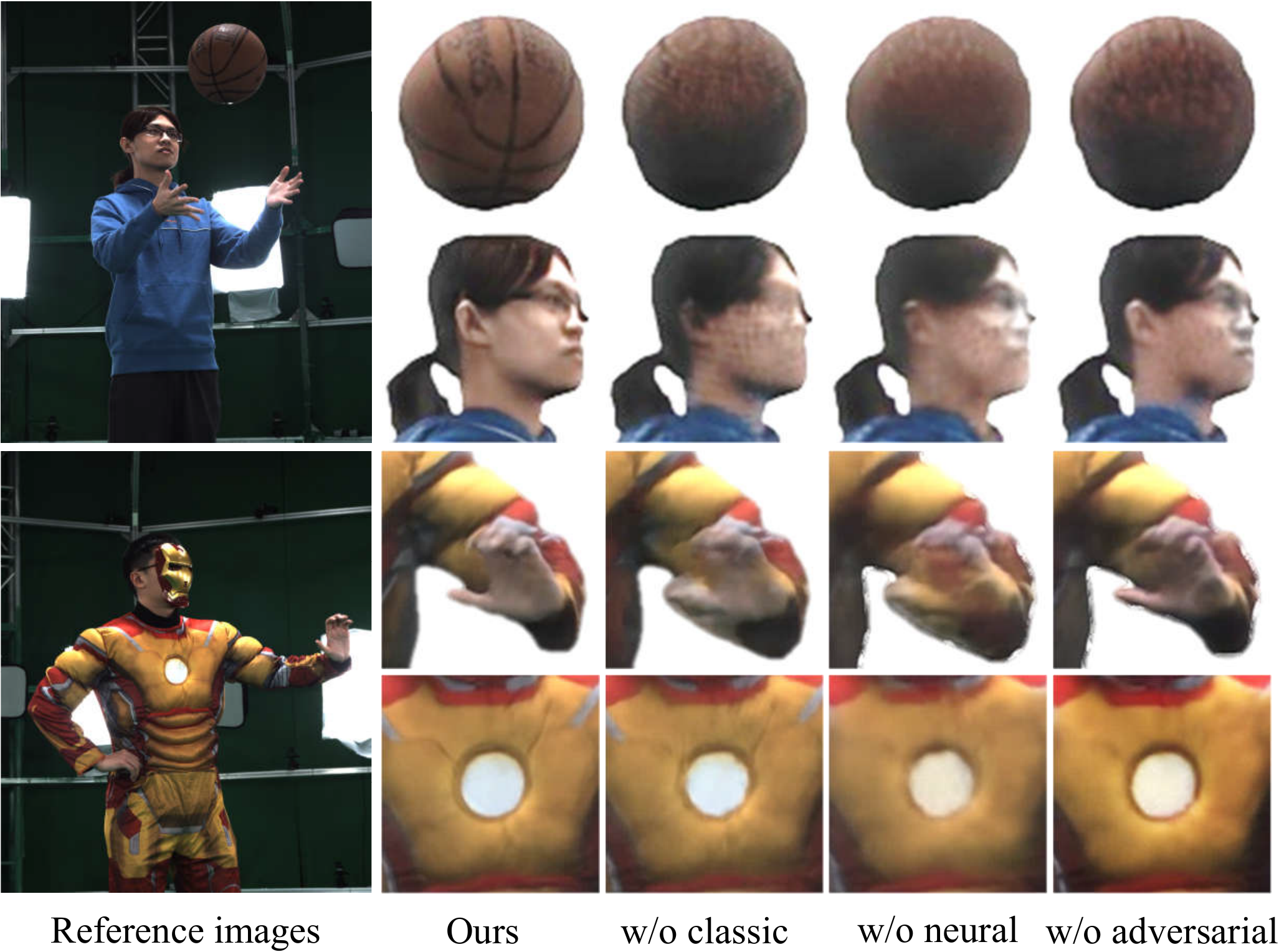}
	\caption{Ablation study for the various components of our approach. Our full pipeline achieves more realistic rendering results.}
	\label{fig:Evaluations}
\end{figure}

Here, we first evaluate the individual components of the proposed FNHR.
Let \textbf{w/o classic} and \textbf{w/o neural} denote the variations of FNHR without the graphic texturing branch and the neural point rendering branch in Sec.~\ref{sec:dualfusion}, respectively.
We also compare against the variation which is supervised directly using the L1 loss and perception loss without the patch-based adversarial training, denoted as \textbf{w/o adversarial}.
As shown in Fig.~\ref{fig:Evaluations}, only a single-branch rendering suffers from uncanny texture artifacts in novel views, while the lack of 
adversarial training leads to severe blur texturing output.
In contrast, our full pipeline enables photo-realistic texture reconstruction in novel views.
For further analysis of the individual components, we utilize the same four metrics as the previous subsection, as shown in Tab.~\ref{tab:Evaluations}. 
This not only highlights the contribution of each algorithmic component but also illustrates that our approach can robustly render texture details.

\begin{figure}[t]
	\centering
	\includegraphics[width=\linewidth]{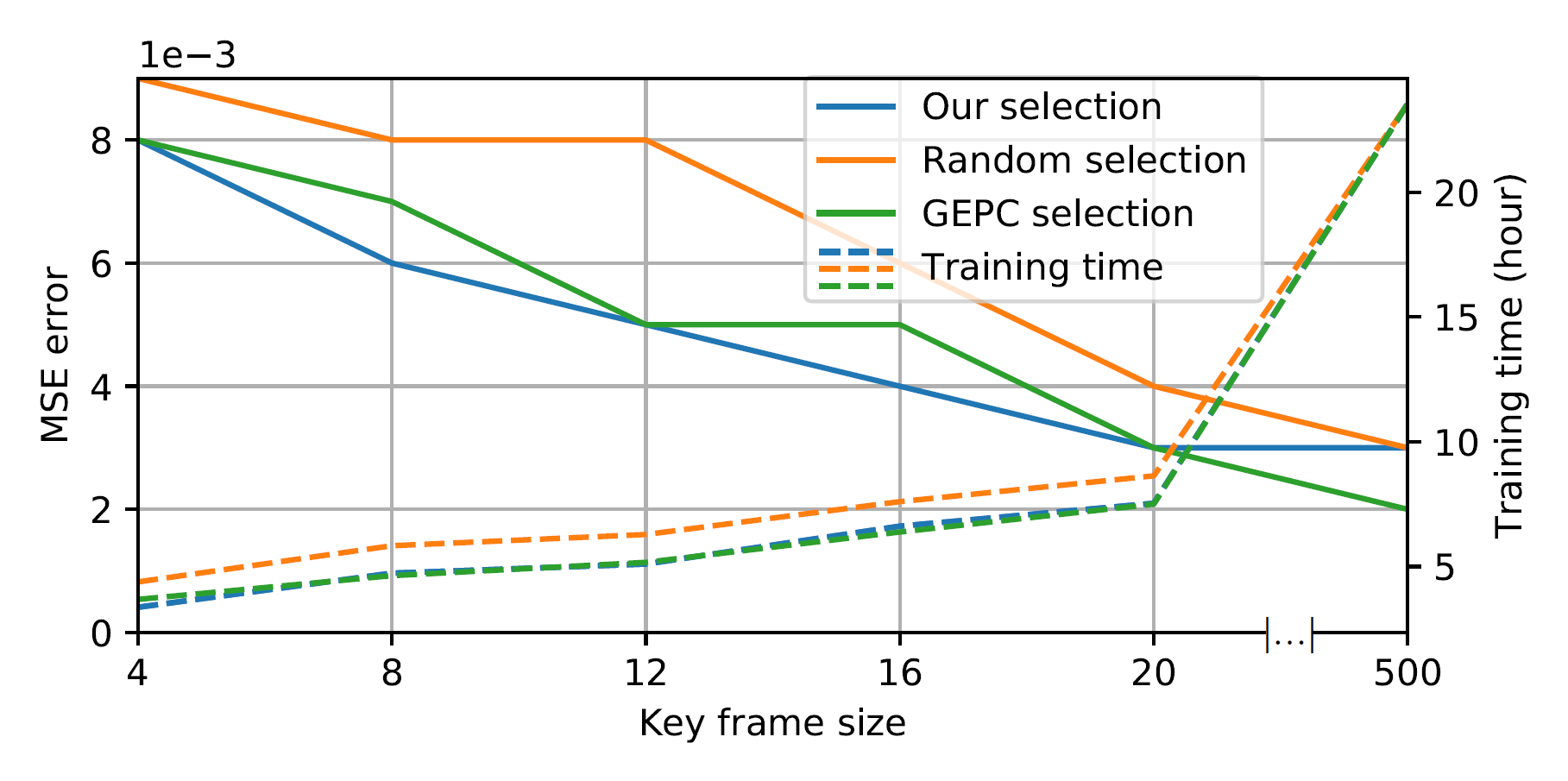}

	\caption{Evaluation of our key-frame selection. Our pose-guided selection scheme consistently outperforms the random one.}

	\label{fig:keyframe}
\end{figure}

We further evaluate our key-frame selection strategy under various numbers of key-frames.
As shown in Fig.~\ref{fig:keyframe}, our pose-guided strategy consistently outperforms the random one in terms of accuracy and training efficiency. The Graph Embedded Pose Clustering (GEPC) method ~\cite{markovitz2020graph} performs sililar to our strategy.
Besides, it can be seen from Fig.~\ref{fig:keyframe} that using more key-frames will increase the training time but improve the accuracy.
Empirically, the setting with 20 key-frames for a sequence with about 500 frames serves as a good compromising settlement between effectiveness and efficiency, which achieves reasonable MSE error and reduces nearly $70\%$ training time.

\begin{table}[t]
	\begin{center}
		\centering
		\resizebox{0.48\textwidth}{!}{
			\begin{tabular}{l|cccc}
				\hline
				Method      & PSNR$\uparrow$ & SSIM$\uparrow$ & Photometric error $\downarrow$ & MSE $\downarrow$ \\
				\hline
				Neural Volumes    & 25.69  & 0.9145  & 0.056 & 0.010 \\
				LookinGood         & 27.11  & 0.9518  & 0.052 & 0.008 \\
				NeRF        & 28.35 & 0.9657  &0.041&0.006  \\
				NHR & 30.08 & 0.9626 &0.038 & 0.004\\
				Ours        & \textbf{33.83}  & \textbf{0.9807} & \textbf{0.025} &\textbf{ 0.002}  \\
				\hline
			\end{tabular}
		}
		\caption{Quantitative comparison against various methods under various metrics. Our method achieve consistently better results.}
		\label{tab:Comparison}
	\end{center}
\end{table}

\begin{table}[t]
	\begin{center}
		\centering

		\resizebox{0.48\textwidth}{!}{
			\begin{tabular}{l|cccc}
				\hline
				Method      & PSNR$\uparrow$ & SSIM$\uparrow$ & Photometric error $\downarrow$ & MSE $\downarrow$ \\
				\hline
				w/o classic        & 28.72 & 0.9107 & 0.039 & 0.005   \\
				w/o neural         & 27.71  & 0.8743  & 0.048 & 0.006  \\
				w/o adversarial & 30.56 &  0.9218 & 0.035 & 0.003\\
				Ours        & \textbf{33.82}  & \textbf{0.9602} & \textbf{0.025} &\textbf{ 0.002}  \\
				\hline
			\end{tabular}
		}
		\caption{Quantitative evaluation for our rendering modules. The result shows that adopt all the branch can generate better result than using either. And with the well designed final renderer the result achieves less errors}
		\label{tab:Evaluations}
	\end{center}
\end{table}

\subsection{Limitation and Discussion} 
We have demonstrated compelling neural rendering results of a variety of challenging scenarios.
Nevertheless, as the first trial of a few-shot neural human performance rendering approach, the proposed FNHR is subject to some limitations.
First, the training process of FNHR to provide a visually pleasant rendering result for about 500 frames takes about 6 to 8
hours, which is not suitable for online applications. 
%
%
Besides, our approach is fragile to challenging self-occlusion motions and severe segmentation error, leading to body-part missing during rendering.
How to handle such challenging scenarios remains an open problem for future research.
Furthermore, our approach relies on a consistent lighting assumption. It's promising to handle complex changing lighting conditions for view-dependent rendering under our challenging sparse setting. 
From the system aspect, our FNHR still relies on six RGBD images to provide a good appearance prior. 
It's an interesting direction to use fewer RGB inputs for data-driven in-the-wild rendering.

    \section{Conclusion}
We have presented the first few-shot neural human performance rendering approach to generate photo-realistic textures of human activities in novel views with only spatially and temporally sparse RGBD training inputs.
%
%
Our experimental results show the effectiveness of our approach for challenging human performance rendering with various poses, clothing types, and topology changes under sparse setting.
We believe that it is critical step for neural human performance analysis, with many potential applications in VR/AR like gaming, entertainment and immersive telepresence.

\end{CJK}
\section*{Acknowledgements}
This work was supported by NSFC programs (61976138, 61977047), the National Key Research and Development Program (2018YFB2100500), STCSM (2015F0203-000-06) and SHMEC (2019-01-07-00-01-E00003)
\appendix

\small
\bibliographystyle{named}
\bibliography{ijcai21}

\begin{thebibliography}{}

\bibitem[\protect\citeauthoryear{Aliev \bgroup \em et al.\egroup
  }{2019}]{aliev2019neural}
Kara-Ali Aliev, Artem Sevastopolsky, Maria Kolos, Dmitry Ulyanov, and Victor
  Lempitsky.
\newblock Neural point-based graphics.
\newblock {\em arXiv preprint arXiv:1906.08240}, 2019.

\bibitem[\protect\citeauthoryear{Cao \bgroup \em et al.\egroup
  }{2017}]{OpenPose}
Zhe Cao, Tomas Simon, Shih-En Wei, and Yaser Sheikh.
\newblock Realtime multi-person 2d pose estimation using part affinity fields.
\newblock In {\em Computer Vision and Pattern Recognition (CVPR)}, 2017.

\bibitem[\protect\citeauthoryear{Chen \bgroup \em et al.\egroup
  }{2019}]{chen2019tightcap}
Xin Chen, Anqi Pang, Yang Wei, Lan Xui, and Jingyi Yu.
\newblock Tightcap: 3d human shape capture with clothing tightness.
\newblock {\em arXiv preprint arXiv:1904.02601}, 2019.

\bibitem[\protect\citeauthoryear{Chen \bgroup \em et al.\egroup
  }{2021}]{chen2021sportscap}
Xin Chen, Anqi Pang, Wei Yang, Yuexin Ma, Lan Xu, and Jingyi Yu.
\newblock Sportscap: Monocular 3d human motion capture and fine-grained
  understanding in challenging sports videos.
\newblock {\em arXiv preprint arXiv:2104.11452}, 2021.

\bibitem[\protect\citeauthoryear{Collet \bgroup \em et al.\egroup
  }{2015}]{collet2015high}
Alvaro Collet, Ming Chuang, Pat Sweeney, Don Gillett, Dennis Evseev, David
  Calabrese, Hugues Hoppe, Adam Kirk, and Steve Sullivan.
\newblock High-quality streamable free-viewpoint video.
\newblock {\em ACM Transactions on Graphics (TOG)}, 34(4):69, 2015.

\bibitem[\protect\citeauthoryear{Dou \bgroup \em et al.\egroup
  }{2017}]{motion2fusion}
Mingsong Dou, Philip Davidson, Sean~Ryan Fanello, Sameh Khamis, Adarsh Kowdle,
  Christoph Rhemann, Vladimir Tankovich, and Shahram Izadi.
\newblock Motion2fusion: Real-time volumetric performance capture.
\newblock {\em ACM Trans. Graph.}, 36(6):246:1--246:16, November 2017.

\bibitem[\protect\citeauthoryear{Joo \bgroup \em et al.\egroup
  }{2018}]{TotalCapture}
Hanbyul Joo, Tomas Simon, and Yaser Sheikh.
\newblock Total capture: A 3d deformation model for tracking faces, hands, and
  bodies.
\newblock In {\em The IEEE Conference on Computer Vision and Pattern
  Recognition (CVPR)}, June 2018.

\bibitem[\protect\citeauthoryear{Li \bgroup \em et al.\egroup
  }{2020}]{li2020neural}
Zhengqi Li, Simon Niklaus, Noah Snavely, and Oliver Wang.
\newblock Neural scene flow fields for space-time view synthesis of dynamic
  scenes.
\newblock {\em arXiv preprint arXiv:2011.13084}, 2020.

\bibitem[\protect\citeauthoryear{Lombardi \bgroup \em et al.\egroup
  }{2019a}]{NeuralVolumes}
Stephen Lombardi, Tomas Simon, Jason Saragih, Gabriel Schwartz, Andreas
  Lehrmann, and Yaser Sheikh.
\newblock Neural volumes: Learning dynamic renderable volumes from images.
\newblock {\em ACM Trans. Graph.}, 38(4), July 2019.

\bibitem[\protect\citeauthoryear{Lombardi \bgroup \em et al.\egroup
  }{2019b}]{lombardi2019neural}
Stephen Lombardi, Tomas Simon, Jason Saragih, Gabriel Schwartz, Andreas
  Lehrmann, and Yaser Sheikh.
\newblock Neural volumes: Learning dynamic renderable volumes from images.
\newblock {\em arXiv preprint arXiv:1906.07751}, 2019.

\bibitem[\protect\citeauthoryear{Markovitz \bgroup \em et al.\egroup
  }{2020}]{markovitz2020graph}
Amir Markovitz, Gilad Sharir, Itamar Friedman, Lihi Zelnik-Manor, and Shai
  Avidan.
\newblock Graph embedded pose clustering for anomaly detection.
\newblock In {\em Proceedings of the IEEE/CVF Conference on Computer Vision and
  Pattern Recognition}, pages 10539--10547, 2020.

\bibitem[\protect\citeauthoryear{Martin-Brualla \bgroup \em et al.\egroup
  }{2018}]{LookinGood}
Ricardo Martin-Brualla, Rohit Pandey, Shuoran Yang, Pavel Pidlypenskyi,
  Jonathan Taylor, Julien Valentin, Sameh Khamis, Philip Davidson, Anastasia
  Tkach, Peter Lincoln, and et~al.
\newblock Lookingood: Enhancing performance capture with real-time neural
  re-rendering.
\newblock {\em ACM Trans. Graph.}, 37(6), December 2018.

\bibitem[\protect\citeauthoryear{Mildenhall \bgroup \em et al.\egroup
  }{2020}]{nerf}
Ben Mildenhall, Pratul~P. Srinivasan, Matthew Tancik, Jonathan~T. Barron, Ravi
  Ramamoorthi, and Ren Ng.
\newblock Nerf: Representing scenes as neural radiance fields for view
  synthesis.
\newblock In Andrea Vedaldi, Horst Bischof, Thomas Brox, and Jan-Michael Frahm,
  editors, {\em Computer Vision -- ECCV 2020}, pages 405--421, Cham, 2020.
  Springer International Publishing.

\bibitem[\protect\citeauthoryear{Mustafa \bgroup \em et al.\egroup
  }{2016}]{mustafa20164d}
Armin Mustafa, Hansung Kim, and Adrian Hilton.
\newblock 4d match trees for non-rigid surface alignment.
\newblock In {\em European Conference on Computer Vision}, pages 213--229.
  Springer, 2016.

\bibitem[\protect\citeauthoryear{Park \bgroup \em et al.\egroup
  }{2019}]{park2019deepsdf}
Jeong~Joon Park, Peter Florence, Julian Straub, Richard Newcombe, and Steven
  Lovegrove.
\newblock Deepsdf: Learning continuous signed distance functions for shape
  representation.
\newblock In {\em Proceedings of the IEEE Conference on Computer Vision and
  Pattern Recognition}, pages 165--174, 2019.

\bibitem[\protect\citeauthoryear{Park \bgroup \em et al.\egroup
  }{2020}]{park2020deformable}
Keunhong Park, Utkarsh Sinha, Jonathan~T Barron, Sofien Bouaziz, Dan~B Goldman,
  Steven~M Seitz, and Ricardo-Martin Brualla.
\newblock Deformable neural radiance fields.
\newblock {\em arXiv preprint arXiv:2011.12948}, 2020.

\bibitem[\protect\citeauthoryear{Ronneberger \bgroup \em et al.\egroup
  }{2015}]{ronneberger2015u}
Olaf Ronneberger, Philipp Fischer, and Thomas Brox.
\newblock U-net: Convolutional networks for biomedical image segmentation.
\newblock In {\em International Conference on Medical image computing and
  computer-assisted intervention}, pages 234--241. Springer, 2015.

\bibitem[\protect\citeauthoryear{Shysheya \bgroup \em et al.\egroup
  }{2019}]{shysheya2019textured}
Aliaksandra Shysheya, Egor Zakharov, Kara-Ali Aliev, Renat Bashirov, Egor
  Burkov, Karim Iskakov, Aleksei Ivakhnenko, Yury Malkov, Igor Pasechnik,
  Dmitry Ulyanov, et~al.
\newblock Textured neural avatars.
\newblock In {\em Proceedings of the IEEE Conference on Computer Vision and
  Pattern Recognition}, pages 2387--2397, 2019.

\bibitem[\protect\citeauthoryear{Su \bgroup \em et al.\egroup
  }{2020}]{robustfusion}
Zhuo Su, Lan Xu, Zerong Zheng, Tao Yu, Yebin Liu, and Lu~Fang.
\newblock Robustfusion: Human volumetric capture with data-driven visual cues
  using a rgbd camera.
\newblock In Andrea Vedaldi, Horst Bischof, Thomas Brox, and Jan-Michael Frahm,
  editors, {\em Computer Vision -- ECCV 2020}, pages 246--264, Cham, 2020.
  Springer International Publishing.

\bibitem[\protect\citeauthoryear{Suo \bgroup \em et al.\egroup
  }{2021}]{suo2021neuralhumanfvv}
Xin Suo, Yuheng Jiang, Pei Lin, Yingliang Zhang, Kaiwen Guo, Minye Wu, and Lan
  Xu.
\newblock Neuralhumanfvv: Real-time neural volumetric human performance
  rendering using rgb cameras.
\newblock {\em arXiv preprint arXiv:2103.07700}, 2021.

\bibitem[\protect\citeauthoryear{Thies \bgroup \em et al.\egroup
  }{2019}]{thies2019deferred}
Justus Thies, Michael Zollh{\"o}fer, and Matthias Nie{\ss}ner.
\newblock Deferred neural rendering: Image synthesis using neural textures.
\newblock {\em ACM Transactions on Graphics (TOG)}, 38(4):1--12, 2019.

\bibitem[\protect\citeauthoryear{Tretschk \bgroup \em et al.\egroup
  }{2020}]{tretschk2020non}
Edgar Tretschk, Ayush Tewari, Vladislav Golyanik, Michael Zollh{\"o}fer,
  Christoph Lassner, and Christian Theobalt.
\newblock Non-rigid neural radiance fields: Reconstruction and novel view
  synthesis of a deforming scene from monocular video.
\newblock {\em arXiv preprint arXiv:2012.12247}, 2020.

\bibitem[\protect\citeauthoryear{Wu \bgroup \em et al.\egroup
  }{2020}]{Wu_2020_CVPR}
Minye Wu, Yuehao Wang, Qiang Hu, and Jingyi Yu.
\newblock Multi-view neural human rendering.
\newblock In {\em Proceedings of the IEEE/CVF Conference on Computer Vision and
  Pattern Recognition (CVPR)}, June 2020.

\bibitem[\protect\citeauthoryear{Xian \bgroup \em et al.\egroup
  }{2020}]{xian2020space}
Wenqi Xian, Jia-Bin Huang, Johannes Kopf, and Changil Kim.
\newblock Space-time neural irradiance fields for free-viewpoint video.
\newblock {\em arXiv preprint arXiv:2011.12950}, 2020.

\bibitem[\protect\citeauthoryear{Xiang \bgroup \em et al.\egroup
  }{2019}]{Xiang_2019_CVPR}
Donglai Xiang, Hanbyul Joo, and Yaser Sheikh.
\newblock Monocular total capture: Posing face, body, and hands in the wild.
\newblock In {\em The IEEE Conference on Computer Vision and Pattern
  Recognition (CVPR)}, June 2019.

\bibitem[\protect\citeauthoryear{{Xu} \bgroup \em et al.\egroup
  }{2019}]{UnstructureLan}
L.~{Xu}, Z.~{Su}, L.~{Han}, T.~{Yu}, Y.~{Liu}, and L.~{FANG}.
\newblock Unstructuredfusion: Realtime 4d geometry and texture reconstruction
  using commercialrgbd cameras.
\newblock {\em IEEE Transactions on Pattern Analysis and Machine Intelligence},
  pages 1--1, 2019.

\end{thebibliography}

\end{document}